\documentclass[letterpaper, 10 pt, conference]{ieeeconf}  

\IEEEoverridecommandlockouts                              

\usepackage{graphics} 
\usepackage{times} 
\usepackage{amsmath} 
\usepackage{amssymb}  
\usepackage{amsfonts}
\usepackage{subfigure}
\usepackage{hyperref}
\usepackage{graphicx}
\usepackage{float}
\usepackage{caption}
\usepackage{multirow}
\usepackage[ruled,linesnumbered]{algorithm2e}
\usepackage{color}
\hypersetup{hidelinks}

\newcommand{\eg}{\textit{e}.\textit{g}., }

\def\ralhigh#1{\textcolor{black}{#1}}

\title{\LARGE \bf
DiPGrasp: Parallel Local Searching for Efficient Differentiable Grasp Planning
}

\author{Wenqiang Xu$^{*1}$, Jieyi Zhang$^{*1}$, Tutian Tang$^{1}$, Zhenjun Yu$^{1}$, Yutong Li$^{1}$ and Cewu Lu$^{1}$
\thanks{*Equal contribution}
\thanks{$^{1}${\tt\small \{vinjohn, yi\_eagle, tttang, jeffson-yu, davidliyutong, lucewu\}@sjtu.edu.cn}. Wenqiang Xu, Jieyi Zhang, Tutian Tang, Zhenjun Yu, Yutong Li are with the School of Electronic Information and Electrical Engineering, Shanghai Jiao Tong University, Shanghai, China. Cewu Lu is the corresponding author, a member of Qing Yuan Research Institute and MoE Key Lab of Artificial Intelligence, AI Institute, Shanghai Jiao Tong University, Shanghai, China.}%
}

\begin{document}

\maketitle
\thispagestyle{empty}
\pagestyle{empty}

\begin{abstract}
Grasp planning is an important task for robotic manipulation. Though it is a richly studied area, a standalone, fast, and differentiable grasp planner that can work with robot grippers of different DOFs has not been reported. In this work, we present DiPGrasp, a grasp planner that satisfies all these goals. DiPGrasp takes a force-closure geometric surface matching grasp quality metric. It adopts a gradient-based optimization scheme on the metric, which also considers parallel sampling and collision handling. This not only drastically accelerates the grasp search process over the object surface but also makes it differentiable. We apply DiPGrasp to three applications, namely grasp dataset construction, mask-conditioned planning, and pose refinement. For dataset generation, as a standalone planner, DiPGrasp has clear advantages over speed and quality compared with several classic planners. For mask-conditioned planning, it can turn a 3D perception model into a 3D grasp detection model instantly. As a pose refiner, it can optimize the coarse grasp prediction from the neural network, as well as the neural network parameters. Finally, we conduct real-world experiments with the Barrett hand and Schunk SVH 5-finger hand.
Video and supplementary materials can be viewed on our website: \url{https://dipgrasp.robotflow.ai}.

\end{abstract}

\section{Introduction}
Dexterous grasping is a long-standing problem in the robotics community. It is a task that achieves object grasp planning with high-DOF multi-finger robot grippers. Compared with the richly studied parallel-jaw grippers \cite{dexnet1,dexnet4,graspnet,6dofgrasp}, dexterous robot hands can perform more complex grasping \cite{grasp_type}, \eg human-like grasp.
However, searching for a proper grasp pose in high-DOF configuration space is not as simple as the parallel-jaw grippers, since the latter only needs to consider the relative pose from the gripper wrist towards the object, while the former needs to determine the finger joint configurations.

Research on dexterous grasp planning lasts for decades \cite{forceclosure,eigengrasp,isf,diffforceclosure}. Methodologies followed by previous researchers can be roughly categorized into two main classes, \textit{analytical} and \textit{data-driven}. The analytical methods \cite{forceclosure,eigengrasp,isf,diffforceclosure} usually follow the model-based path and search for a grasp pose that can meet the requirements of some certain grasp quality metrics \cite{forceclosure,isotropy_metric}. Previous works on this track are generally slow, with a typical speed of 15s $\sim$ 20min to generate one valid dexterous grasp pose. On the other hand, data-driven methods leverage learning algorithms like deep learning \cite{ffhnet,twostage,ddg} and reinforcement learning \cite{pixel_attentive,rl_affordance} to predict grasp pose from noisy input of unseen objects. Once the network is properly trained, the inference time of grasp generation can be lowered to $30$ms \cite{ffhnet}. Methods on this track require a large amount of training data, which take considerable time (\eg 7 hours for 10K dexterous grasps in \cite{dexgraspnet}) to generate.

\begin{figure}[t!]
    \centering
    \includegraphics[width=0.99\linewidth]{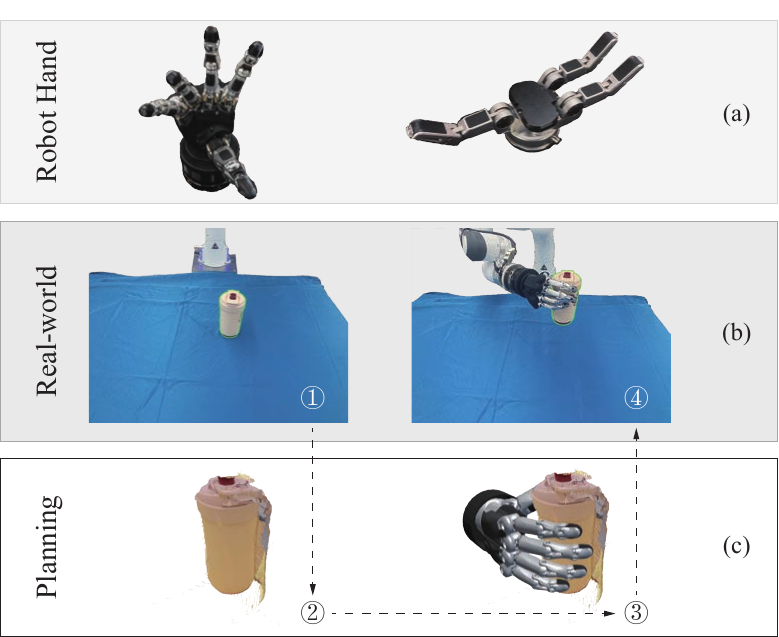}
    \caption{DiPGrasp can (a) work with robot grippers with different DOFs. (b-c) It can produce high-DOF grasp poses efficiently from the observed point cloud and guide the execution in the real world. }
    \label{fig:fig1}
    \vspace{-10pt}
\end{figure}

Based on these observations, we determine a practical grasp planner should take the legacy of the conventional analytical path, but also can support the research on data-driven approaches. Thus, it should be \textit{standalone}, \textit{fast}, and \textit{differentiable}. As a \textit{standalone} planner, it can produce valid grasps based on a certain grasp metric and work with arbitrary grippers. As a \textit{fast} planner, it can generate as many valid grasps as quickly as possible. As a \textit{differentiable} planner, it takes gradient-based optimization techniques to solve the planning problem and can work with neural networks. To meet all these goals, we present \textbf{DiPGrasp}.

\begin{figure*}[t!]
    \centering
    \includegraphics[width=0.90\linewidth]{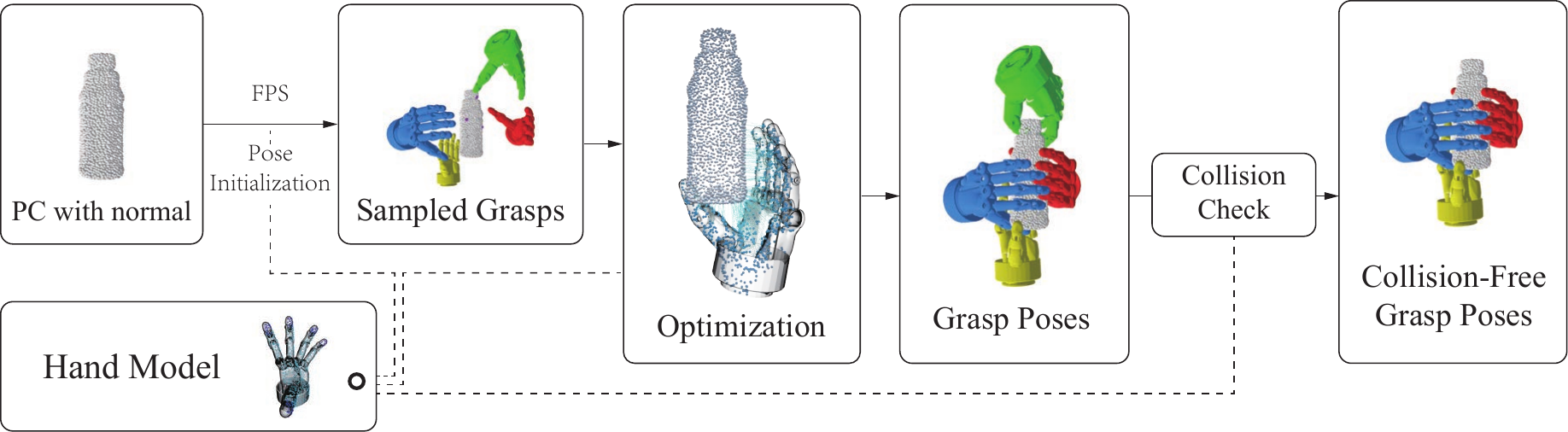}
    \caption{DiPGrasp pipeline. DiPGrasp takes a point cloud with normal as input. It first samples locations on the point cloud (red dot) and initializes the pose accordingly. Then it operates the differentiable optimization process to generate the grasps.}
    \label{fig:pipeline}
\end{figure*}

DiPGrasp is inspired by a geometry-based surface matching metric from prior research \cite{isf}. We add a force-based regularization term to enhance the grasp stability, and propose a novel force-based surface matching metric.
To search the optimal grasps under this metric, DiPGrasp adopts the sample-optimize approach, initiating with sampled poses and refining them using the metric's objective function. Given the quadratic nature of the proposed force-based surface matching metric, it is amenable to gradient descent. By utilizing differentiable computing techniques \cite{pytorch}, we efficiently batch sample poses and use gradient-based optimization like stochastic gradient descent (SGD). Both batch processing and gradient descent optimizers are commonplace in contemporary learning platforms \cite{pytorch}. This method greatly expedites the grasp search, allowing for simultaneous optimization at \textbf{all sampled locations}. In practice, DiPGrasp can perform a grasp search with a Schunk SVH hand using $80$ initial poses in just $2.5$s on an NVIDIA GeForce RTX 3080 GPU, using 8 GB memory, averaging $\sim118$ms per valid grasp. This is substantially faster than EigenGrasp \cite{eigengrasp}, which takes $\sim20$s for a single grasp.

To prove the efficacy of the proposed DiPGrasp, we have applied it to three different applications: \textit{Grasp dataset construction}, \textit{Mask-conditioned planning}, and \textit{Pose refinement}. Using DiPGrasp, we could construct a large dexterous hand dataset faster than the SOTA method \cite{dexgraspnet} with a much higher valid proportion, generate valid grasp poses just on partial point cloud mask, or improve the quality of coarse poses generated by a neural network.

We summarize our contributions as follows:
\begin{itemize}
    \item We introduce DiPGrasp, a differentiable, fast grasp planner compatible with robot grippers of varied DOFs. It employs gradient-based optimization for local grasp pose searches and can operate in parallel. This differentiability allows seamless integration into any differentiable frameworks.
    \item We use DiPGrasp for grasp dataset construction, mask-conditioned planning, and pose refinement. Real-world robot tests are conducted on mask-conditioned planning using models trained from the grasp dataset.
\end{itemize}

\section{Related Work}
\subsection{Analytical Dexterous Grasp Planner}
Dexterous grasp planning research often employs heuristic metrics to assess grasp quality using a known object model, maximizing these metrics to find good grasp poses \cite{forceclosure,isotropy_metric}. Yet, real-world scenarios frequently involve objects with incomplete or unknown models. The high-DOF configuration space search for optimal grasp poses is non-convex, making it challenging to locate the best solution. Consequently, optimization methods like simulated annealing \cite{eigengrasp}, quadratic programming \cite{isf}, and Bayesian optimization \cite{diffforceclosure} are introduced, though they can be computationally time-consuming.

Regarding differentiable grasp planning, Liu et al. \cite{diffgrasp_rss} introduce a differentiable $\epsilon$-metric addressed with semidefinite programming, but it necessitates significant adjustments to serve as a standalone planner. The dataset for their study is sourced from GraspIt! \cite{graspit}. Another approach by Liu et al. \cite{diffforceclosure} employs a gradient-based method for grasp pose generation, but it is not very efficient as it yields limited successful grasps after lengthy computations. Grasp'D \cite{graspd} offers a grasp synthesis process relying on differentiable physics simulation, but its results are sequential and demand a full object mesh. Contrarily, DiPGrasp can explore grasp poses concurrently and works with partial point clouds which could be generated from RGB-D observations instead of full meshes (See Fig. \ref{fig:fig1}).

\subsection{Data-driven Dexterous Grasp Planner}
Data-driven grasp planning is increasingly recognized, with the capability to process incomplete object information, such as partial point clouds \cite{ffhnet}, RGB \cite{rl_affordance}, RGB-D images \cite{pixel_attentive,ddg}, and volumes \cite{twostage}. These techniques, often underpinned by deep or reinforcement learning, can generalize to unknown objects with partial information.

For those employing deep neural networks, the common practice is either a generative \cite{ffhnet,twostage,ddg} or discriminative \cite{unigrasp,ddg} task to predict grasps. However, given neural networks' prediction limitation in precision, several methods \cite{dvgg,voxel1} use a refinement step for improved accuracy. Some first propose indirect representations like heatmaps \cite{heatmap} or probabilistic distributions \cite{ddg} from which grasp poses are derived.

Incorporating reinforcement learning for dexterous grasping is challenging due to the complexity of dexterous configurations. Wu et al. \cite{pixel_attentive} employ attention mechanisms for policy learning, while Mandikal and Grauman \cite{rl_affordance} utilize affordance data, eventually learning from videos with human actions \cite{rl_video}.

\section{Force-based Surface Matching Metric}\label{sec:isf}
Previous works \cite{isf,fan2019optimization} consider grasp planning as a surface-matching problem between the robot gripper and the object to be grasped. Such formulation can induce an intuitive optimization objective. However, it overlooks an important aspect of grasping, grasping stability. To amend that, we introduce a force-closure term in the optimization objective functions. In the following, we will first describe the geometry-based surface matching objectives, and then describe how to add the force-closure regularization into it.

\subsection{Geometry-based Surface Matching}
After a grasp is achieved, there will be multiple contacts between the gripper and object surfaces.
Each contact $i$ is defined by $(\mathcal{S}^f_i,\mathcal{S}^o_i)$, where $\mathcal{S}^f_i$ is the finger contact surface, and $\mathcal{S}^o_i$ is the object contact surface. $\mathcal{S}^f_i$ is a subset of $i$-th finger link surface $\partial \mathcal{F}_i$. $\partial \mathcal{F}_i$ is transformed by $\mathcal{T}(\cdot)$ given gripper pose $\mathcal{P}=(R, t, q)$, where $R \in SO(3)$, $t \in \mathbb{R}^3$ are 6-DOF robot gripper wrist pose, and $q \in \mathbb{R}^k$ is the $k$-DOF finger pose. $q_i$ is bounded by $q_{\text{min,i}}$ and $q_{\text{max,i}}$, which are the joint limits respectively. On the other hand, the object contact surface $\mathcal{S}^o_i$ is the nearest neighbor (NN) of $\mathcal{S}^f_i$ on the object surface $\partial\mathcal{O}$.

Then, we can formulate the grasp planning problem by searching the optimal grasp pose $\mathcal{P}$ by minimizing the surface alignment error $E$:

\begin{align}
   &\min_{R, t, q} \sum_{i=1}^k E(\mathcal{S}^f_i,\mathcal{S}^o_i)\\
   s.t. \quad & \mathcal{S}^f_i \in \mathcal{T}(\partial \mathcal{F}_i; R,t,q)\\
   & \mathcal{S}^o_i = NN_{\partial\mathcal{O}}(\mathcal{S}^f_i)\\
   & q_{min, i} \leq q_{i} \leq q_{max, i} \\
   & i=1,\ldots,k 
\end{align}

The objective function can be reformed with two terms: point matching error $E_p$ and normal alignment error $E_n$.
\begin{equation}
   E_{sm}(R, t, q)=E_p(R, t, q) + E_n(R),
\end{equation}
\begin{equation}\label{equ:ep}
   E_p(R,t,q)=\sum_{i=1}^k\sum_{j=1}^m ||(x_{j_i}-y_{j_i})^Tn^y_{j_i}||^2_2,
\end{equation}
\begin{equation}\label{equ:en}
   E_n(R)=\sum_{j=1}^m ||(Rn_{j_i}^x)^Tn^y_{j_i}+1||^2_2.
\end{equation}
$E_p$ measures the point-to-plane error between the $j$-th point on finger link $i$, $x_{j_i}$ and the matched point $y_{j_i}$ on object, $n^y_{j_i}$ is the normal vector at point $y_{j_i}$. $x_{j_i} \in S^f_i$, and thus is related to $(R,t,q)$. $E_n$ encourages the normals of the finger surface to align towards the normals of the object surface.
A more detailed formulation description can be referred to \cite{isf}.

\subsection{Force-based Surface Matching}
The surface-matching heuristics is intuitive and easy to optimize. However, it does not consider the force stability, thus the optimization objectives are inclined to generate an ``in-contact'' configuration rather than an ``in-grasp'' configuration.

To amend this, we introduce a variant of the force-closure term from \cite{diffforceclosure}, it can work with object models in point cloud form and a differentiable optimization scheme. We modify the point matching error in Eq. \ref{equ:ep} to:

\begin{equation}\label{equ:efp}
   E_{fp}(R,t,q)=E_p(R,t,q) +||Gc||_2,
\end{equation}
 and
\begin{align}
G & =\left[\begin{array}{cccc}
I_{3 \times 3} & I_{3 \times 3} & \ldots & I_{3 \times 3} \\
\left\lfloor x_1\right\rfloor_{\times} & \left\lfloor x_2\right\rfloor_{\times} & \ldots & \left\lfloor x_n\right\rfloor_{\times}
\end{array}\right], \\
\left\lfloor x_i\right\rfloor_{\times} & =\left[\begin{array}{ccc}
0 & -x_i^{(3)} & x_i^{(2)} \\
x_i^{(3)} & 0 & -x_i^{(1)} \\
-x_i^{(2)} & x_i^{(1)} & 0
\end{array}\right] .
\end{align}
 where $x_i$ represents the contact point on hand and $c_i$ denotes the corresponding normal vector of $x_i$. We utilize Farthest Point Sampling (FPS) to discern contact points, specifically by selecting four points $(n=4)$ from the palm side of the hand that fall within the top 20\% of points nearest to the object.
For the details of the force closure term please refer to \cite{diffforceclosure}.

Finally, we give the force-based surface matching heuristics:
\begin{equation}\label{equ:e}
    E(R, t, q)=E_{fp}(R, t, q) + E_n(R).
\end{equation}
With this metric, we can design a grasp planner for arbitrary objects and robot hands.

\section{DiPGrasp}\label{sec:sfddp}
In this section, based on the force-based surface matching metric, we first describe how to optimize the objective function and make the process differentiable in Sec. \ref{sec:reduce_optimizetion}. A collision-aware term to the grasp quality metrics $E$ is introduced in Sec. \ref{sec:collision}. And the parallel sampling strategy in Sec. \ref{sec:sample}. Finally, we will describe an adjustable weighting map for different grasp type priors in Sec. \ref{sec:weight}.

The pseudocode of fully differentiable DiPGrasp is given in Algo. \ref{alg:diffisp}. The overall pipeline is illustrated in Fig. \ref{fig:pipeline}.

\begin{algorithm}[t!]
\caption{DiPGrasp}
\label{alg:diffisp}
\KwIn{Initial state $R_0$, $t_0$, $q_0$, object surface $\partial\mathcal{O}$, gripper surface $\partial\mathcal{F}$, error threshold $\epsilon_0$, max iterations $N_1$, $N_2$}
\KwOut{$\hat{R}^*$, $\hat{t}^*$, $\hat{q}^*$}
\BlankLine
$i_1 \leftarrow 0$, $i_2 \leftarrow 0$\;
$Loc \leftarrow sample(\partial O)$; \tcp{Sec. \ref{sec:sample}}
$(R_s, t_s, q_s) \leftarrow initial\_pose(Loc)$; \tcp{Sec. \ref{sec:sample}}
Collision check; \tcp{Sec. \ref{sec:collision}}
\While{$\Delta\epsilon \geq \epsilon_0$ and $i_1 \leq N_1$}{
    $S^f_{i_1} \leftarrow \mathcal{T}(\partial\mathcal{F}\otimes \mathcal{W}; R_s, t_s, q_s)$; \tcp{Sec. \ref{sec:weight}}
    $S^o_{i_1} \leftarrow NN_{\partial \mathcal{O}}(S^f_i)$\;
    \While{$\Delta\epsilon \geq \epsilon_0$ and $i_2 \leq N_2$}{\tcp{Sec. \ref{sec:reduce_optimizetion}}
        $\mathcal{L}\leftarrow E^*(S^f_{i_1}, S^o_{i_1})$\;
        $dR, dt, dq \leftarrow \frac{\partial \mathcal{L}}{\partial R}, \frac{\partial \mathcal{L}}{\partial t}, \frac{\partial \mathcal{L}}{\partial q}$\;
        $R\leftarrow R-\alpha dR$\;
        $t\leftarrow t-\beta dt$\;
        $q\leftarrow q-\gamma dq$\;
        $\Delta\epsilon\leftarrow |E^*-E_{prev}|$\;
    } 
}
Collision check (Sec. \ref{sec:collision}); \tcp{Sec. \ref{sec:collision}}
$\hat{R}^*, \hat{t}^*, \hat{q}^* \leftarrow R, t, q$\;
\end{algorithm}

\subsection{Optimizing Grasp Pose with A Gradient-based Solver}\label{sec:reduce_optimizetion}

In this work, we directly use gradient descent for both wrist and finger pose optimization. Since $R,t,q$ are the parameters to update, and as we can see from Algo. \ref{alg:diffisp}, all the operations (Line 6, 9-13) relevant to these parameters are differentiable. We can preserve the gradients for these parameters during iterations so that the gradient can be used to update the parameters of itself or a differentiable pipeline.

Gradient-based optimization is known to suffer from local minima \cite{isf}. But modern gradient descent-based optimizers (e.g., SGD \cite{sgd}) usually adopt momentum or even second-order momentum (e.g., Adam \cite{adam}) to try to escape from the local minima. Thus, though previous gradient-based works \cite{isf} have proposed many complex optimization schemes, we find the commonly used SGD is stable enough.

\subsection{Collision Handling}\label{sec:collision}
In the original surface matching heuristics, $E$ does not consider avoiding collision during optimization. This will cause a large portion of the planned grasp pose to be in the collision and result in a failed grasp. Thus, in order to save more collision-free grasp poses, collision handling is desired.

In this section, we will introduce a differentiable barrier term $E_b$ and a fast collision check method to avoid collisions.

\paragraph{Barrier Term}
We first consider a distance measure between two points $d_i = (x_{j_i} - y_{j_i})^T(x_{j_i} - y_{j_i})$, and a barrier boundary  $\hat{d}$. If two points are getting too close, the energy between them will grow exponentially. No repulsion will be applied if $d_i \geq \hat{d}$. Besides, we would like the barrier term to have at least $C^1$ continuity for gradient computation. With these ideas in mind, we borrow the definition of barrier term from \cite{ipc}:

\begin{equation}
    E_{b} = \frac{1}{m}\left \{ 
    \begin{aligned}
        &(d_i - \hat{d})^2\ln{(\frac{d_i}{\hat{d}})},&0 < d_i < \hat{d} \\
        &0. & d_i \geq \hat{d}
    \end{aligned} \right.
\end{equation}
This barrier function has $C^2$ continuity, which is sufficient for gradient descent. 
We set $\hat{d}=0.05$ for all experiments.

To note, the barrier term can also be used to prevent joints from updating to an out-of-range configuration. We can define $d_{q_{min,i}}=|q_i-q_{min_i}|$ and $d_{q_{max,i}}=|q_i-q_{max_i}|$ to barrier the joint to get close to the limits. $\hat{d}_{min}=\hat{d}_{max}=(d_{max}-d_{min})\times0.15$. These terms give $E_{b, q_{min}}$ and $E_{b, q_{max}}$.

We add $E_b$ and $E_{b, q}$ to $E$ and result in $E^*$
\begin{equation}
    E^*(R, t, q) = E(R, t, q) + E_b(R, t, q) + E_{b, q}(R,t,q).
\end{equation}

\paragraph{Collision Check}
Though the barrier term can pull away the points from getting too close, it does not guarantee intersection-free results. Thus, after the optimization process, a collision check is essential for a better result. Since collision detection \cite{ipc} is computationally expensive, here we adopt a simple collision detection approach.

\begin{figure}[t!]
    \begin{minipage}[b]{.49\linewidth}
    \centering
    \subfigure[]{\label{fig:collision}\includegraphics[width=1\linewidth]{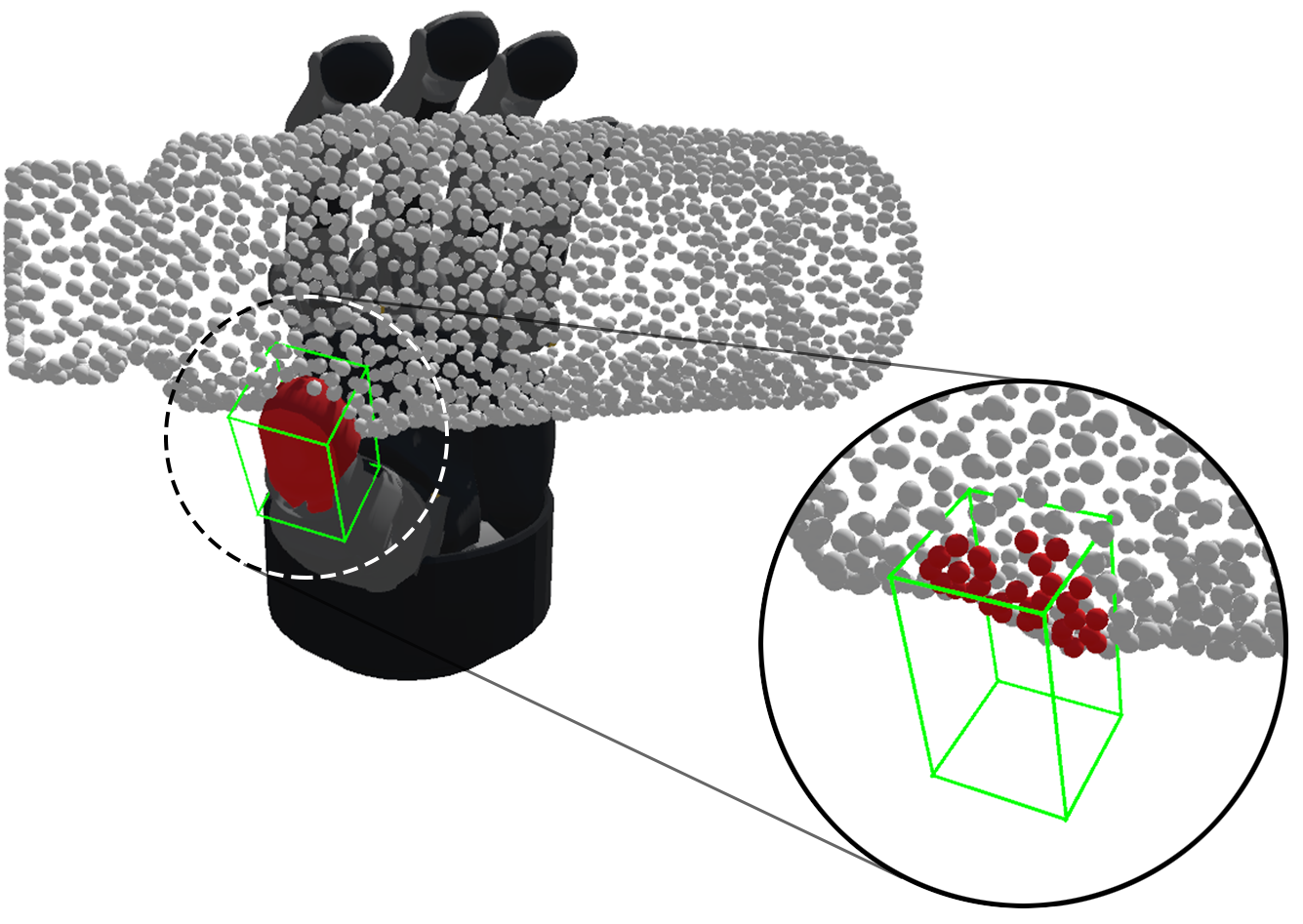}}
    \end{minipage}
    \begin{minipage}[b]{.49\linewidth}
    \subfigure[]{ \label{fig:hand-model}\includegraphics[width=1\linewidth]{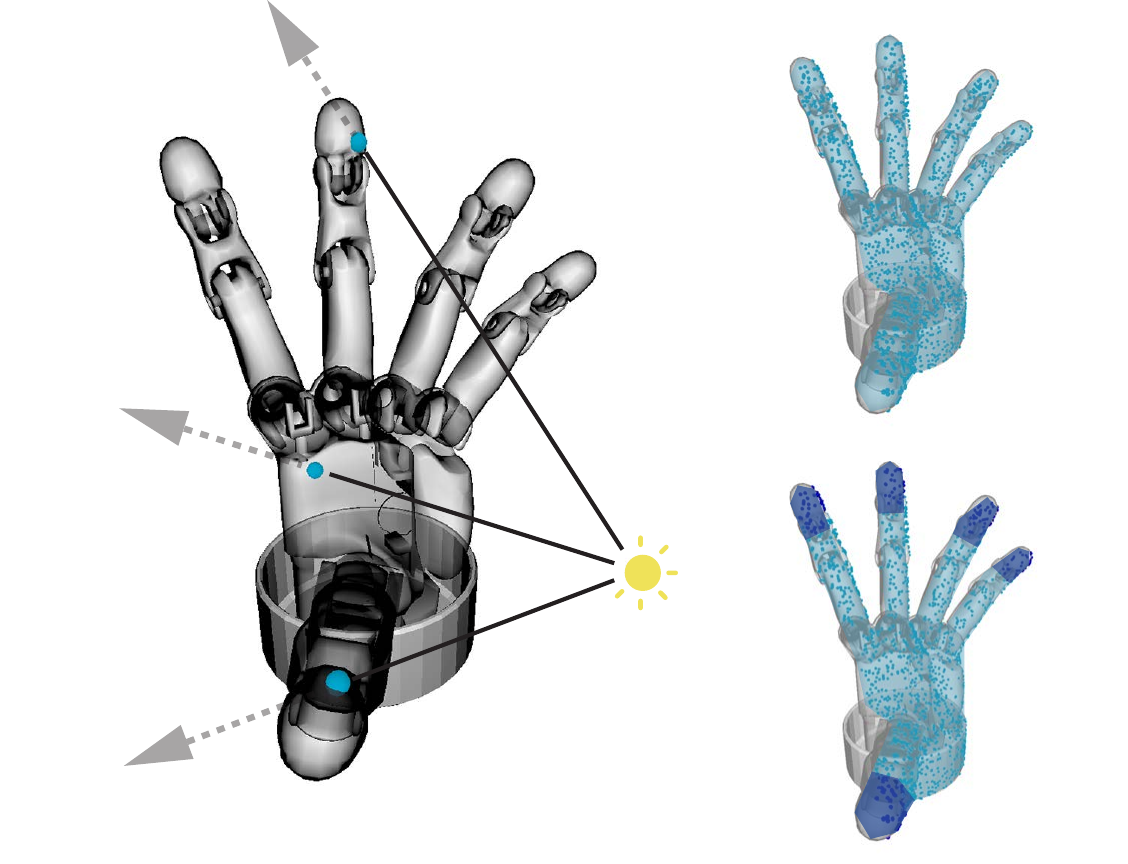}}
    \end{minipage}
    \caption{(a) Collision check. Some points on the object surface are in collision (in red) with the bounding box, which represents the fingertip. (b) \textbf{Left}: Gripper weighting map can be automatically generated by ray casting. \textbf{Right}: Finger links like fingertips can easily be singled out according to the kinematic structure. (Darker area means bigger weight)}
\end{figure}

As shown in Fig. \ref{fig:collision}, we represent each finger link and the palm as a bounding box. Then, we can tell if there are potential collisions by checking if any point is inside the bounding box. And we also apply this method to detect if self-collision happens. This step can be easily computed in parallel.

\subsection{Parallel Sampling}\label{sec:sample}
In this section, we will describe the sampling strategy.

As shown in Algo. \ref{alg:diffisp}, the gradient-based solver reacts to one location at a time. Thus, we can utilize the ``batchify'' techniques which are commonly adopted in modern deep learning training \cite{pytorch} for parallel sampling and optimization. By sampling a bunch of initial poses and organizing them into a batch, we can search for valid grasp poses from these initial poses simultaneously. In this way, we can search $K$ poses at the same runtime as 1 initial pose. The maximum of $K$ is limited by the GPU memory and will be discussed in Sec. \ref{sec:exp}.

For each sampled pose, we make the palm directly oriented to the sampled points, spread the finger, and place it at a distance of $d_{gripper}$ from the point. $d_{gripper}$ is different according to the grippers, as the finger lengths of different grippers may be different. For Barrett hand $d_{gripper} = 15$cm, for Schunk SVH hand $d_{gripper} = 12$cm.

We will first do a collision check to make sure the hand and object are not in collision with the initial pose.

The extensive sampling can ensure multiple valid grasp outputs and the coverage of the object surface. We may have different strategies to select the one valid grasp to execute the plan. We will describe the grasp selection we use in Sec. \ref{sec:exp}.
Besides, we can also limit the sampling according to a certain constraint, e.g. mask on the surface. For example, an object mask from a scene, or an affordance mask from an object. We will give examples for the object mask in the application of \textbf{Mask-conditioned Planning}.

\subsection{Gripper Weighting Map}\label{sec:weight}

It is obvious that we only want to align the object surface with the palmar side surface to form a grasp. \ralhigh{If the dorsal side gets involved in the matching between the hand and object, the object might be aligned to the dorsal side, or trapped between the dorsal and palmar sides of the hand.} However, there's no off-the-shelf algorithm that could separate the palmar side surface given a gripper (or a hand) model. One way is to define the palmar side surface of the gripper manually, which is laborious for new grippers. 

Here, we describe a simple yet effective approach to annotating the palmar area of any grippers. As shown in Fig. \ref{fig:hand-model} \textbf{Left}, inspired by the ray-casting pipeline \cite{raycasting}, we place a light source in the palm, then cast large quantities of rays omnidirectionally. Then, we regard all the first-intersection points on the gripper surface for each ray as the palmar side points. Empirically, the location of the light source can be given by calculating the average location of all the fingertips. The whole process can be made automatically in this way.

Besides, we introduce a weight map to separate the different parts of the surface, which could make the grasp pose more diverse. For example, we could assign a bigger weight to the fingertip part while assigning a smaller to the other side, as shown in Fig. \ref{fig:hand-model} \textbf{Right}. The fingertip is prone to getting closer to the surface, leading to a pinch grasp.

\begin{figure*}[t!]
    \centering
    \includegraphics[width=0.8\linewidth]{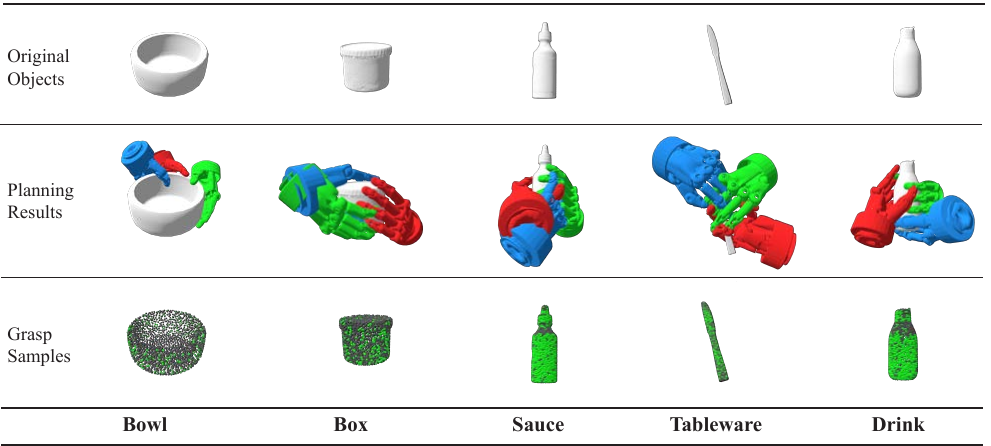}
    \caption{Grasp planning results. \textbf{Upper}: Original object models. \textbf{Middle}: Sampled grasp planning results for visualization. The grasp poses in the display are randomly selected from the valid grasp poses in the \textbf{Bottom} row.   
    \textbf{Bottom}: Distribution of the grasp planning results. Green means a valid grasp is found towards this point. Black means no valid grasp found.}
    \label{fig:dataset1}
\end{figure*}

\section{Experiments}\label{sec:exp}
\subsection{Task Definition}
We set three tasks to show the efficiency and utility of DiPGrasp, namely \textbf{Grasp Dataset Construction}, \textbf{Mask-conditioned Planning}, and \textbf{Pose Refinement}. 

\paragraph{Grasp Dataset Construction} \ralhigh{We construct a comprehensive dataset for grasp analysis using RFUniverse \cite{rfuniverse} and plan to apply it across multiple applications. Our dataset includes 50 object models across five categories—bowl, box, sauce, tableware, and drink bottle—sourced from the AKB-48 dataset \cite{akb48}. We select 8 models from each category for training and 2 for testing.}

\ralhigh{For robot grippers, we utilize the Barrett hand ($7$-DOF) and the Schunk SVH 5-finger hand ($20$-DOF). Our system can be compatible with various grippers, provided they have a URDF file.}

\ralhigh{To get the grasp poses for each object, we initially sample $2000$ points on each object's surface using farthest point sampling (FPS), and apply DiPGrasp to determine potential grasp locations. These poses are later validated by RFUniverse \cite{rfuniverse}. A successful grasp was defined by the gripper's ability to lift and hold the object $20$cm above its start position for $3$ seconds, even when interfered with randomly altered gravitational forces for $1$ second.}

\ralhigh{Further, we construct scenes by randomly placing five grasp-annotated objects on a table. The grasp poses are transformed according to the object poses. We filter out any poses in collision with other objects. Each scene is captured using simulated RGB-D cameras, which employ IR-based depth rendering to minimize the sim-to-real depth discrepancy, aiding in training neural networks.}

\ralhigh{In total, our dataset comprises $2000$ scenes, with $1800$ designated for training and $200$ for testing. For object grasp pose generation, the time efficiency of grasp searches varied between the Barrett and Schunk SVH hands, with the former requiring $25$ minutes and the latter $50$ minutes to process all $50$ models.} \ralhigh{In terms of scene construction, each scene takes about $1$ minute to generate. This process includes placing objects and filtering collided grasps, which could also be executed in parallel. Overall, the Schunk SVH hand yields $1.2$ million collision-free grasps ($1$ million for training, $0.2$ million for testing), while the Barrett hand produces 2.8 million ($2.2$ million for training, $0.6$ million for testing).}

\paragraph{Mask-conditioned Grasp Planning} \ralhigh{Given a scene point cloud containing objects to grasp as input, we first use an off-the-shelf 3D point cloud instance segmentation algorithm to separate each instance point cloud, outputting instance masks. The training data are generated by taking RGB-D snapshots from three views in the scene, which then are merged into a scene point cloud. To mimic the real-world multi-camera calibration error, we augment the scene point cloud when merging multi-view RGB-D by giving them a random positional offset around $1$cm.} 

\ralhigh{To train Mask3D, we adopt the original hyper-parameters from its official implementation. Learning rate is $0.0001$. Batch size is $4$. The training epoch is $400$.}
\ralhigh{For each mask predicted by Mask3D, we use FPS to sample $50$ points for pose initialization, and use DiPGrasp to generate grasp poses.}
\paragraph{Pose Refinement} For each object instance mask, we could also use a neural network to generate grasp poses. We modify a PointNet++ \cite{pointnet++} (called SimpleGrasp) to generate point-wise coarse grasp poses. We change the output layer of the PointNet++ to generate a (1+7+k)-d vector. 1-D for validness of the point (whether it is a good point to grasp). 7-d for the wrist pose which is represented by a 4-d quaternion vector and 3-d translation. k-d for the joint angle. Then, we could use the DiPGrasp to refine the coarse poses concurrently.

\ralhigh{To train PointNet++, we use the exact hyper-parameters in its official implementation. Learning rate is $0.0001$. Batch size is $8$. The training epoch is $200$. The optimizer is SGD with a momentum of $0.9$.}

The Grasp Dataset Construction task is designed to demonstrate the utility of DipGrasp as a standalone grasp planner. All three tasks reflect different procedures in a learning-based grasp detection pipeline.

\subsection{Metrics}
\paragraph{$\epsilon$-metric} $\epsilon$-metric \cite{forceclosure} is widely adopted to measure the force closure quality.
\paragraph{Barrier-augmented Surface Matching (BSM) metric} Our planner is based on the barrier-augmented surface matching metric $E^*$. It can also be used to measure the grasp quality.
\paragraph{Valid proportion} As our method can give multiple outputs simultaneously, we can examine the valid proportion for all the initial positions. The valid proportion is defined by the number of valid grasps over the number of initial positions. A valid grasp is a grasp pose generated when the optimization is converged, which does not necessarily guarantee successful execution in simulation or the real world. It is an important metric for dataset generation and coverage evaluation. 
\paragraph{Success rate} The success rate is defined by the number of successful grasps in simulation over the number of valid grasps generated. It takes into account the grasp selection and execution. We test the planned grasps in RFUniverse \cite{rfuniverse} and the real world to see the success rate. 

\subsection{Experiment results}
\subsubsection{Grasp Dataset Construction} \label{sec:exp_grasp_dataset}
For Barrett hand, over all categories, we can generate a grasp dataset with a $67.66\%$ valid proportion after physics-based simulation. For Schunk SVH hand, we have a $26.5\%$ valid proportion after physics-based simulation.

The reasons for the low valid proportion on the object surface are threefold: (1) It is natural that not every location is suitable for grasping. (2) The grasp search process is highly non-convex, making local minima inevitable. (3) Thin and small objects like tableware are hard to grasp stably in simulation.

However, since we have $2000$ initial sample locations to search over the object surface, we still have over $1300$ valid grasps for Barrett hand, and over $500$ valid grasps for Schunk hand. 
Besides, as searching at $80$ locations will consume 8GB GPU memory, we have to split the search $25$ times with our NVIDIA GeForce RTX 3080 GPU. Totally, it takes $40$s for Barrett hand and $60$s for Schunk hand to accomplish searching at $2000$ locations. The average time for a valid grasp is $30$ms and $118$ms respectively. Even faster speed can be achieved with GPUs of larger memory, \eg RTX 3090 and A100. 

We compare our method with previous methods regarding the quality and speed in Table \ref{tab:data_con}.
\begin{table}[ht!]
    \centering
    \begin{tabular}{c|ccc}
        \hline
        Method & $\epsilon$ & valid proportion (\%) &  ave. time (s) \\\hline
        ISF (B) \cite{isf} &0.159 & 61.9 & 2.13 \\
        EigenGrasp (B) \cite{eigengrasp} & 0.237 & / &18.75 \\
        DexGraspNet*  \cite{dexgraspnet} & 0.1145 & 18 & 2.47 \\\hline
        Ours (B) & 0.201 & 67.66 & 0.030 \\
        Ours (S) &0.170 & 26.50 & 0.118 \\\hline
    \end{tabular}
    \caption{* is reported for reference, the dataset they generate has similar object shapes to ours, and the speed is measured with A100 GPU. We only compare ISF and EigenGrasp with Barrett (B) because the implementations of these methods do not support Schunk hand (S). EigenGrasp can only produce 1 grasp pose at a time, thus it does not have a valid proportion. We report the time it takes to produce the first valid grasp, and the time is measured on the same computational hardware setting.}
    \label{tab:data_con}
\end{table}

\subsubsection{Performance in Mask-conditioned Planning}\label{sec:exp_mask}
We use Mask3D \cite{mask3d} as the segmentation module. For each instance mask, we get several valid grasp poses after the collision check. In simulation, we can test every valid grasp pose. But in the real world, it is time-consuming to verify every grasp, thus we select the grasp pose from the valid grasps with the minimum $E^*$, and check the success rate.

We only carry out the grasp planning for segmented objects with a confidence prediction larger than $0.7$. The scores in Table \ref{tab:mask_result} are based on these detections. We compare our method with baseline methods ISF \cite{isf} and the grasp generation algorithm used in DexGraspNet \cite{dexgraspnet} on real-world tests, and report the success rate in Table \ref{tab:mask_result}. For the success rate of our method evaluated in the simulated scenes, please refer to the supplementary materials.
\begin{table}[ht!]
    \centering
    \begin{tabular}{c|c|c|c}
        \hline
        Env & tries & success & success rate\\\hline
        ISF (Barrett) & $50$ & $20$ & $40\%$\\
        DexGraspNet (Schunk) & $50$ & $11$ & $22\%$\\ \hline
        Ours (Barrett) without force& $50$ & $26$ & $52\%$ \\
        Ours (Schunk) without force& $50$ & $22$ & $44\%$\\
        Ours (Barrett) & $50$ & $30$ & $60\%$\\
        Ours (Schunk) & $50$ & $24$ & $48\%$\\
        \hline
    \end{tabular}
    \caption{Success rate for mask-conditioned Planning with different planning algorithms and grippers in real worlds.}
    \label{tab:mask_result}
\end{table}

\subsubsection{Performance in Pose Refinement}\label{sec:exp_refine}
For pose refinement, we train SimpleGrasp with the segmented object point clouds and adopt the valid grasps generated by DiPGrasp as the corresponding grasp pose label. It is a common setting in grasp detection task \cite{graspnet}.
\begin{figure}[ht!]
    \centering
    \includegraphics[width=0.99\linewidth]{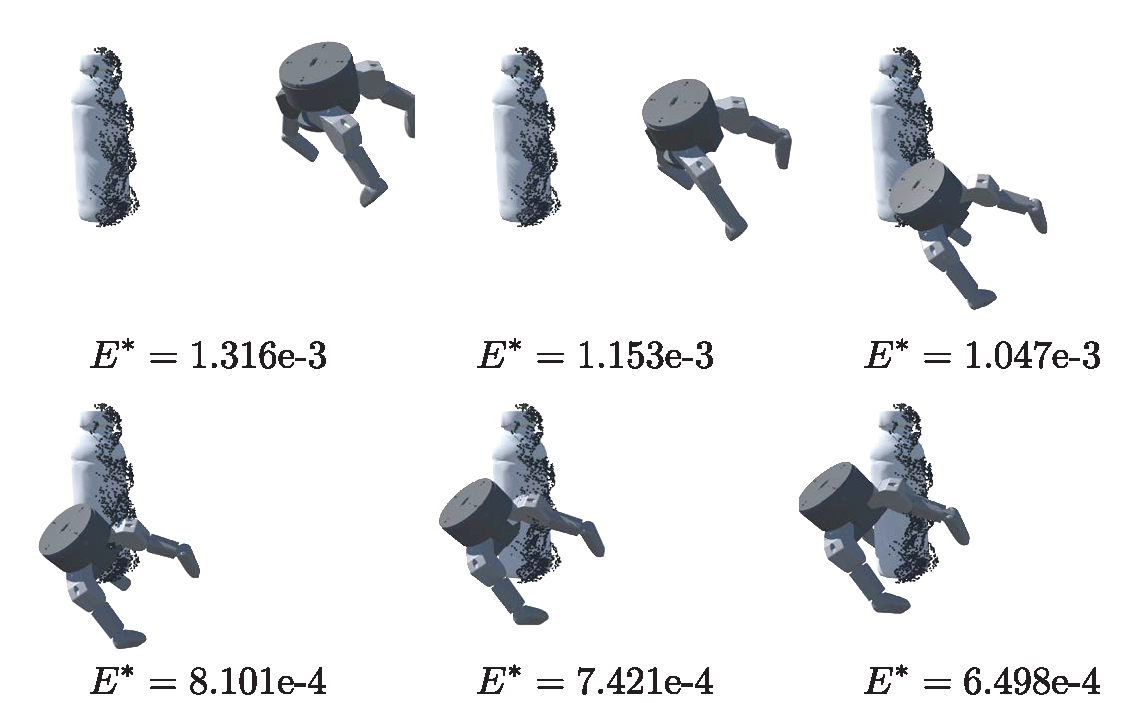}
    \caption{Pose refinement. Even though the neural network prediction is extremely coarse, we still can progressively make it a better grasp with DiPGrasp.}
    \label{fig:pose_refiner}
\end{figure}
\paragraph{Neural Network Refinement}
Using gradient back-propagating from $E^*$, our differentiable planner can help optimize the parameters in SimpleGrasp neural networks. As our planner can be optimized with a gradient descent-based optimizer, they can be jointly trained with the same SGD optimizer. We train SimpleGrasp alone first, and then jointly train with DiPGrasp for the final $40$ epochs. In comparison with training without DiPGrasp, incorporating DiPGrasp can improve the coarse grasp $\epsilon$-metric from $0.085$ to $0.101$.

To note, DiPGrasp needs to join in the late stage of training, as there are few good initial poses for DiPGrasp to carry on in the early stage. 

\paragraph{Pose Refinement} Apparently, though neural networks can predict good grasp poses sometimes, further refinement is needed in most cases. Our planner can make it with ease. In table \ref{tab:pose_refine}, we should assume no object model knowledge during the test stage. Therefore, we cannot calculate the $\epsilon$-metric, and report the BSM-metric instead. Besides, we do not collect valid grasps, as the optimization process has not yet been completed. Thus, we report the total runtime (speed) instead of the average speed. A qualitative result is illustrated in Fig. \ref{fig:pose_refiner} with Barrett hand.
\begin{table}[t!]
    \centering
    \begin{tabular}{ccc}
        \hline
        Method & BSM-metric ($\times 10^{-5}$) & speed (ms) \\\hline
        \textit{Barrett} \\
        SimpleGrasp & 863.61 & 132\\  
        3-step &  148.44 & 332\\ 
        6-step &  90.08 & 532\\
        10-step & 67.43 &  812\\
        13-step & 60.32 & 1022\\\hline
        \textit{Schunk}\\
        SimpleGrasp & 3323.51 &133\\ 
        3-step & 669.42 & 405\\
        6-step & 381.05 & 687\\
        10-step & 266.21 & 1059\\
        13-step & 227.39 & 1314\\\hline
    \end{tabular}
    \caption{DiPGrasp can refine the pose predicted by SimpleGrasp.}
    \label{tab:pose_refine}
\end{table}

\subsection{Discussion}
\subsubsection{Failure Modes}
\ralhigh{Observing the failure examples in real-world experiments, we found that there are three typical failure modes:}
\ralhigh{
\begin{itemize}
    \item The contact happens before the finger reaches the desired configuration, thus the object could be pushed far away. Such a mode often happens on the Barrett because the finger link is too long.
    \item The contact surface is so smooth, it can lead to slip. Such mode often happens on the porcelain bowl with Schunk SVH hand.
    \item The object is so thin that the gripper can not hold it. Such a mode often happens on the tableware.
\end{itemize}
}
\subsubsection{Noise Sensitivity}
\ralhigh{Besides, as a geometry-based surface matching algorithm, it is imperative to evaluate how noise within the point cloud affects performance. To this end, we introduced Gaussian noise to the original point cloud at varying standard deviations and subsequently assessed the impact on algorithmic performance, with the results detailed in Table \ref{tab:noisy_data}. It is evident that the valid proportion declines sharply when the standard deviation reaches 1 cm. Despite this, our algorithm remains sufficiently efficient on data with significant noise levels.}

\begin{table}[ht!]
    \centering
    \begin{tabular}{ccccc}
        \hline
         & no noise & $\sigma$ = 2mm  & $\sigma$ = 5mm & $\sigma$ = 1cm\\\hline
         \textit{Barrett} \\
        Valid Proportion(\%) & 67.66 & 64.32 & 55.98 & 37.45\\
        Ave Time(s) & 0.030 & 0.032 & 0.036 & 0.054 \\ \hline
        \textit{Schunk} \\
        Valid Proportion(\%) & 26.50 & 25.11 & 21.21 & 14.70\\
        Ave Time(s) &  0.118 & 0.125 & 0.147 & 0.213 \\\hline
    \end{tabular}
    \caption{Performance with noised data.}
    \label{tab:noisy_data}
\end{table}
\section{Conclusion and Future Works} 
\label{sec:conclusion}
In this work, we introduce DiPGrasp, a standalone, fast, and differentiable grasp planner compatible with various robot gripper DOFs. DiPGrasp employs force-based surface-matching heuristics for less time-consuming gradient-based optimization and uses parallel computing for simultaneous grasp searches. We've utilized DiPGrasp in grasp dataset construction, mask-conditioned planning, and pose refinement.

As a differentiable planner, in the future, we aim to integrate it with other research areas such as learning to sample, adaptive weight mapping for grasp types, and optimizing objective functions. Additionally, we seek to create a differentiable dexterous manipulation framework using DiPGrasp.






\section*{ACKNOWLEDGMENT}
This work was supported by the National Key Research and Development Project of China (No. 2022ZD0160102) National Key Research and Development Project of China (No. 2021ZD0110704), Shanghai Artificial Intelligence Laboratory, XPLORER PRIZE grants.



\bibliographystyle{IEEEtran}
\bibliography{main}

\end{document}